\documentclass[10pt,twocolumn,letterpaper]{article}

\usepackage[review]{cvpr}      % To produce the REVIEW version
\usepackage{graphicx}
\usepackage{amsmath}
\usepackage{algorithmic}
\usepackage{amssymb}
\usepackage{booktabs}

\usepackage[ruled,norelsize]{algorithm2e} % DO NOT USE FLOATING
\usepackage[pagebackref,breaklinks,colorlinks]{hyperref}

\usepackage[capitalize]{cleveref}
\crefname{section}{Sec.}{Secs.}
\Crefname{section}{Section}{Sections}
\Crefname{table}{Table}{Tables}
\crefname{table}{Tab.}{Tabs.}

\begin{document}

% \title{Shared Affordance-aware Autonomous Manipulation (SAAM)}
\title{Shared Affordance-awareness via Augmented Reality \\ for Proactive Assistance in Human-robot Collaboration}
%\title{Shared affordance-awareness for Proactive Assistance via Augmented Reality \\ in human-robot Collaboration (SPAARC}
%\title{Shared Environment Perception using Augmented Reality for Affordance-aware Human-robot Collaboration} (SEPARAC)

\author{Drake Moore\\
% Northeastern University\\
% Boston, Massachusetts, USA\\
{\tt\small moore.dr@northeastern.edu}
\and
Mark Zolotas\\
{\tt\small m.zolotas@northeastern.edu}
\and
Ta\c{s}k{\i}n~Pad{\i}r\\
{\tt\small t.padir@northeastern.edu}
}
\maketitle

%%%%%%%%% ABSTRACT
% \begin{abstract}
%    Enabling humans and robots to collaborate effectively during a shared task requires extensive communication and knowledge of each others' affordances. Prior work in human-robot collaboration has incorporated knowledge of affordances for assistive scenarios, such as robotic wheelchairs for independent mobility. In this work-in-progress paper, we propose an affordance-based framework involving augmented reality (AR) headsets to highlight the different areas of operation of the human and robot in a shared task environment. Using an AR headset, we can leverage the different perspectives from the robot and human user to inform the robot's autonomous behavior. Moreover, an AR feedback interfaces enables the robot and human to communicate with one another and prompt for assistance when attempting to perform tasks outside their respective affordable area.
% \end{abstract}

\begin{abstract}
   Enabling humans and robots to collaborate effectively requires purposeful communication and an understanding of each other's affordances. Prior work in human-robot collaboration has incorporated knowledge of human affordances, i.e., their action possibilities in the current context, into autonomous robot decision-making. This ``affordance-awareness'' is especially promising for service robots that need to know when and how to assist a person that cannot independently complete a task. However, robots still fall short in performing many common tasks autonomously. In this work-in-progress paper, we propose an augmented reality (AR) framework that bridges the gap in an assistive robot's capabilities by actively engaging with a human through a \textit{shared} affordance-awareness representation. Leveraging the different perspectives from a human wearing an AR headset and a robot's equipped sensors, we can build a perceptual representation of the shared environment and model regions of respective agent affordances. The AR interface can also allow both agents to communicate affordances with one another, as well as prompt for assistance when attempting to perform an action outside their affordance region. This paper presents the main components of the proposed framework and discusses its potential through a domestic cleaning task experiment. %We envision accessible perceptually-enabled shared autonomy in human-robot collaboration in assistive settings.
\end{abstract}

%%%%%%%%% BODY TEXT

\section{Introduction}
\label{sec:intro}

Human-robot collaboration (HRC) is increasingly prevalent in industrial, commercial, and household settings. Nevertheless, HRC remains a naturally complex problem when robots must transition into environments with an unpredictable structure~\cite{abbink_topology_2018}, or when assisting a person with limited physical capabilities due to disability~\cite{demiris_knowing_2009}. In such scenarios, the robot and human agents may encounter areas where they are unable to operate. Whenever an agent is unable to access a desired object or area during a collaborative task, it becomes vital to successful task completion that this state of deadlock is made transparent to both agents~\cite{rosen_mixed_2020,inkulu_challenges_2022}. As a result, each agent in the HRC must not only have knowledge of their own limitations, but should also have sufficient information about their partner's capabilities in order to proactively request assistance if necessary.

\begin{figure}[t]
    \centering
    \includegraphics[scale=0.155]{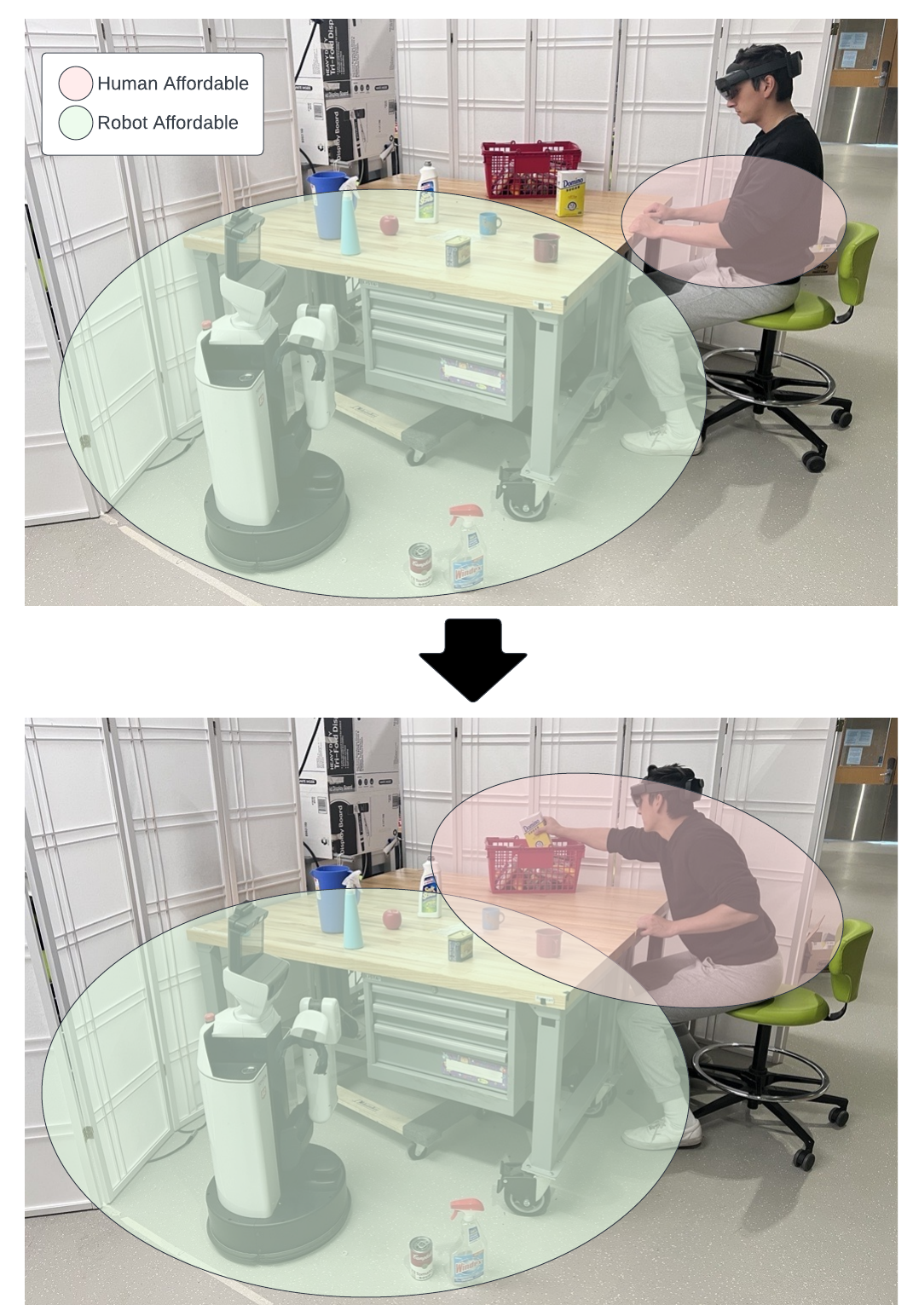}
    \caption{Illustration of human and robot affordances in a collaborative cleaning task. The human's affordable area begins as a small ellipsoid centered around their chest, extending to their hands. As the person grasps and places an object into the allocated bin, their affordable area grows to include novel areas they interacted with.}
    \label{fig:overview}
    \vspace{-4.0mm}
\end{figure}

A prominent means of accurately modeling and conveying agent capabilities within their environment is to represent these capabilities as \textit{object affordances}~\cite{goodrich_seven_2003,koppula_anticipating_2016,jamone_affordances_2018}. Defining object affordances as the possible actions an agent can perform on an object~\cite{gibson_theory_1977,jamone_affordances_2018}, recent works in HRC have utilized knowledge of these affordances to inform the autonomous robot's responsive behavior~\cite{koppula_anticipating_2016,jinpeng_object_2019,quesada_proactive_2022}. However, many of these works have modeled \textit{either} the human or robot's affordances, before then uni-directionally transmitting this information to the other agent. Bi-directional communication of agent intentions, affordances, and spatial perspective continues to be one of the most pressing modern challenges for HRC~\cite{inkulu_challenges_2022}.

% In this work-in-progress paper, we propose an augmented reality (AR) framework to establish ``affordance-awareness'' in \textit{both} agents through a shared spatial understanding of the collaborative task environment. Using RGB-D data available to the AR headset and the robot's sensor suite, we can capture the human and robot's perceptual perspectives of the environment, as well as dynamically construct affordances via agent-specific affordance models. Given these individual environment perspectives and affordance models, a shared spatial representation of affordable areas can be generated. Additionally, the AR headset interface offers the human and robot a means of relaying information on their respective affordance regions to one another. This paper primarily considers object affordances in the form of \textit{reachability}, assuming an object in reach can always be acted on by the agent, as shown in Fig.~\ref{fig:overview}. Human affordable areas can then be mathematically defined based on kinematic models of humans~\cite{lenarcic_simple_1994,iqbal_using_2004}.

In this paper, we propose an augmented reality (AR) framework to establish ``affordance-awareness'' in \textit{both} agents through a shared spatial understanding of the collaborative task environment. Using RGB-D sensors available to the AR headset and robot, the environment is perceived from both agent perspectives. Object affordances are then computed in terms of \textit{reachability} via agent-specific models, where an object in reach is assumed affordable by the agent. We model robot affordances as manipulability ellipsoids that are pre-computed from the robot's known kinematics and a previously mapped voxel grid of the environment~\cite{vahrenkamp_robot_2013,xu_planning_2020,zacharias_capturing_2007}. Human affordances are instead generated based on kinematic models of human reachability~\cite{lenarcic_simple_1994,iqbal_using_2004}, and expressed as a binary voxel grid. These individual agent perspectives and affordances produce a shared spatial representation of affordable areas. Moreover, the AR interface offers the human and robot a means of relaying information on their respective affordance regions to one another. 

Adopting this terminology for affordances, the remainder of this paper introduces the main components of the proposed framework and grounds its applicability in a domestic cleaning task (see Fig.~\ref{fig:overview}). Under this task setup, we demonstrate how each agent can reason over generated affordances to proactively communicate intent, request assistance, and perform collaborative behaviors.

\section{Related Work}
\label{sec:related}

A plethora of research has investigated the application of AR technologies as a communication medium in HRC. For instance, multiple studies have used AR headsets to visualize the robot's internal model and planned trajectories, thereby improving human decision-making during collaboration~\cite{zolotas_head_2018,chakraborti_projection_2018,suzuki_augmented_2022}. Leveraging dual perspectives provided by an AR headset and the robot's vision sensors has also been used to build comprehensive spatial maps of a shared environment to guide agent exploration~\cite{reardon_communicating_2019}. AR interfaces in HRC have additionally been explored to communicate robot intentions for the purpose of improving human safety and trust~\cite{chadalavada_thats_2015,walker_communicating_2018,tsamis_intuitive_2021}. These previous works exhibit uni-directional communication of the robot's state, which benefits solely the human's reasoning.

Bi-directional recognition of internal human and robot states is an auspicious trend in the field of HRC~\cite{inkulu_challenges_2022}, for which AR headsets hold great promise as a bridge of communication~\cite{rosen_mixed_2020}. Recent work has investigated methods of portraying a robot's sensory state to human partners and then enabling the human user to interactively edit the robot's resulting behavior~\cite{chandan_arroch_2021,delmerico_spatial_2022}. Over a bi-directional communication channel, robots may also request assistance from human users via the AR interface~\cite{muhammad_creating_2019}. In these bi-directional communication setups, the role of the human and robot agents becomes even more relevant, specifically in terms of whether they are \textit{passive} or \textit{active}. Bi-directional communication channels where \textit{both} the human and robot could \textit{actively} perceive and manipulate holograms have been shown to yield better task performance~\cite{qiu_human-robot_2020}.

Another essential element of AR headset interfaces relates to \textit{how} information on agent capabilities, e.g., represented as affordances, is presented~\cite{zolotas_head_2018}. Affordances have been modeled in various forms to enhance the quality of HRC, such as object grasp positions, spatial usability, object usage, or reachability~\cite{koppula_anticipating_2016,jamone_affordances_2018,nagarajan_learning_2020}. Object affordances can be modeled for both human and robot manipulation to encode physical constraints in a 2-D semantic map~\cite{fan_vision-based_2022}. Alternatively, the number of agent visits to spatial areas and the type of activity performed at that occupancy cell can be used to encode an affordance representation of an environment~\cite{limosani_long-term_2015,bahl_affordances_2023}. Affordance-aware AR user interfaces for HRC have also been developed to present humans with a semantic understanding of a robot's atomic control actions, but no human-to-robot communication was established~\cite{quesada_proactive_2022}.

%In our paper, we are primarily concerned with the ability of a human agent to reach and act upon a given region. Therefore, we refer to \textit{affordances} as the ability for an agent to reach and act upon given areas and objects within the environment. Our usage of the term affordances partially falls in line with early research into human reachability, in which kinematic models of humans are used to mathematically represent operable areas \cite{lenarcic_simple_1994, iqbal_using_2004}.

%Our work seeks to dynamically model human affordances and leverage the known limitations to inform collaborative behavior. 

While prior work has achieved effective bi-directional communication in HRC, we posit that the advantages obtained from using AR headset interfaces have not yet been fully leveraged. Our major insight is that the \textit{shared} knowledge of each agent's affordances will prove more effective than a one-way communication channel. To achieve shared affordance-awareness, we introduce an AR framework that constructs an environmental representation of affordances from the visual perspectives and estimated intentions of \textit{both} agents in the HRC. A key benefit of this bi-directional information flow is that the human and robot can proactively request assistance from their partner whenever they cannot afford an action. In this paper, the term \textit{affordances} will from hereon refer to the ability for an agent to reach and act upon given areas and objects within the environment.

%\subsection{Mark Reference Notes}
%Benefit of AR head-mounted displays as a feedback interface is the increased sense of presence and engagement over monitors or projectors. For instance, AR headsets have been shown to provide a better overall user experience when disambiguating verbal requests made by a human partner through visualizations overlaid on the shared workspace~\cite{sibirtseva_comparison_2018}. 

\section{Shared Affordance-awareness Framework}

\begin{figure*}
    \centering
    \includegraphics[width=0.95\textwidth]{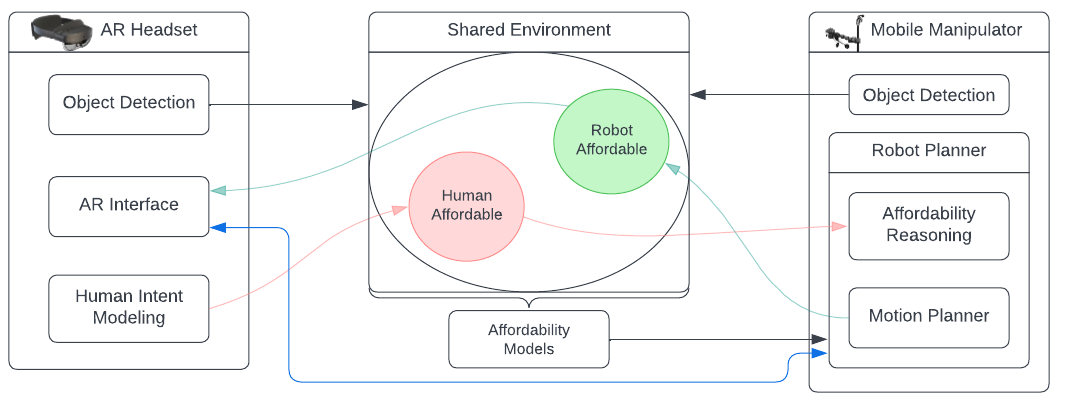}
    \caption{Key components of the augmented reality (AR) framework for affordance-awareness in human-robot collaboration. Perceptual information on objects is obtained through an AR headset and a robot's sensors to create a shared spatial map. The \textit{affordable areas} of the human and robot can then be delineated from agent actions on this shared environment (denoted by red and green circles). Modeled affordances, along with a bi-directional communication channel of human and robot intent, will inform both agent's decision-making. The human and robot can also query each other for assistance (blue line) when objects fall outside their respective affordable areas.}
    \label{fig:architecture}
    \vspace{-1.4mm}
\end{figure*}

In this section, we describe a shared affordance-aware framework for HRC based on an AR headset interface (see Fig.~\ref{fig:architecture} for the architecture overview). The framework combines both the action and perception spaces of the human and robot. Therefore, the human and robot agents are now capable of querying their partner for assistance whenever objects lie outside their own affordable areas. The overall architecture will target the HoloLens 2 headset and a mobile manipulator robot, built atop the ROS 2 middleware~\cite{macenski_ros2_2022}.

\subsection{Intent Communication}
\label{sec:intent}

In a collaborative task, human and robot agents must be able to communicate and understand each other's intent. There are many ways of representing agent intent in HRC~\cite{losey_review_2018,zolotas_disentangled_2022}, but for this work we describe intent as a target object in 3-D Cartesian space that an agent $i$ plans to interact with, $\mathbf{o}_i \in \mathbb{R}^3$. We also only consider a single human and robot agent in the collaboration, ergo $i=\{H,R\}$.

We plan to adopt a similar approach to prior work in communicating robot intent by displaying planned robot trajectories and safety constraints via the user's AR headset~\cite{walker_communicating_2018,tsamis_intuitive_2021}. From the egocentric AR headset's perspective of the human user's hands and eye gaze, we can also predict hand and gaze trajectories to estimate human intent~\cite{huang_using_2015,trick_multimodal_2019}. By presenting the robot with an estimate of the human's grasp plan, the robot can preemptively make adjustments to continue performing safe and collaborative behavior~\cite{lee_chang_effects_2018}. Furthermore, a human can act around the robot's planned behavior by relying on the AR interface's feedback.

\subsection{Spatial Computing}

Spatial computing is the ability for devices to track key features in an environment and represent them digitally~\cite{delmerico_spatial_2022}. Azure Spatial Anchors (ASA) is a mixed reality toolkit for the localization of egocentric devices. This service can be run on both the HoloLens and a mobile robot via the open-source ROS package\footnote{\href{https://github.com/microsoft/azure\_spatial\_anchors\_ros}{https://github.com/microsoft/azure\_spatial\_anchors\_ros}}. Once spatial anchors are determined and shared between the AR headset and robot, a general world frame can be defined for the environment. By simultaneously running ASA on the AR headset and mobile robot, the two sensory sources can be co-localized in a common environment, thus establishing a shared world frame to spatially represent environment features.

In addition to using ASA, we will employ popular object detection and tracking methods to find the locations of objects pertaining to the collaborative task~\cite{wu_detectron2_2019}. For our task, object positions in 3-D space are estimated using the center of the object's bounding box in the image space, coupled with its median filtered depth reading. All object positions can be localized in the shared world frame, which is accessible to both the robot and human.

\subsection{Affordance Modeling}

Modeling affordances is an active multidisciplinary area of research with a diverse array of computational methods~\cite{jamone_affordances_2018}. In this work, we will simplify the problem to treat affordances as spatial regions within an environment, where these regions are delineated by the agent's reachability of objects. Every object in the environment that can be interacted with is assumed ``affordable'' by both the human and robot, such that the affordances, $\mathcal{A}_i$, of an agent $i$ do not need to explicitly model action-object relations~\cite{koppula_anticipating_2016,jamone_affordances_2018}. Instead, $\mathcal{A}_i$ can be represented as a voxel grid of reachable states estimated using the depth sensors of both agents.  

For human affordances, $\mathcal{A}_H$, the voxel grid must capture areas in the environment where a human can operate~\cite{qiu_human-robot_2020, fleet_physically_2014}. In an assistive robotics setting, human agents will often be unwilling or unable to perform actions outside of a given region. As a result, we begin by modeling the entire environment as non-affordable to the human. Throughout the task, we will continuously update the human affordances set, $\mathcal{A}_H$, with new voxels based on the kinematic reachability of human arms, as observed from both the user's and robot's perspectives~\cite{lenarcic_simple_1994, rodriguez_bringing_2003, evangelista_belo_xrgonomics_2021}.

% As a result, we begin modeling the human's affordable area to be constrained to their immediate surroundings, as depicted in the top of Fig.~\ref{fig:overview}. As the human agent performs actions within the environment, we extend the affordable area as a simple ellipsoid extending from the user's chest to the maximum perceived distance of the user's extended hand (see bottom of Fig.~\ref{fig:overview}). The complete human affordable space is a voxel grid that dynamically expands according to ~\cite{evangelista_belo_xrgonomics_2021,rodriguez_bringing_2003}.

To model robot affordances, $\mathcal{A}_R$, we will adopt  existing methods of reasoning over a robot's kinematic constraints given a known spatial map of the environment. These methods represent the capacity for a mobile manipulator to reach and interact with the environment by storing the robot's manipulability ellipsoids at every pose in the discretized workspace~\cite{xu_planning_2020,zacharias_capturing_2007}. We pre-compute these manipulability ellipsoids for the entire environment in an offline manner, resulting in a voxelized representation of the robot's affordable workspace~\cite{vahrenkamp_robot_2013}. While this provides an accurate estimate of the robot's affordable area, we still perform collision-free motion planning for each target object to ensure the system is robust to environment changes. 

\subsection{Proactive Assistance Communication}

By sharing representations of affordances and intentions to both agents in the HRC, we present a framework that best informs the collaborative agent's decision-making. For example, the robot is given its own affordances, the human's affordances, the target object location, and the human's intent, which in turn can be used to guide the autonomous robot motion planning (see Algorithm~\ref{alg:affordance_robot_alg}). A core benefit of using AR headsets in this framework is that both agents are able to request assistance  from their partner via verbal commands or visualizations in the interface. Note how in Algorithm~\ref{alg:affordance_robot_alg}, the robot queries for help from the human agent, e.g., as a graphical cue like a directional arrow~\cite{zolotas_head_2018}, whenever there is no collision-free trajectory, $\tau_R$, towards its target object. By enabling this bi-directional communication channel, human and robot agents can reason about the shared state of the environment and \textit{proactively} query each other for assistance whenever necessary.

\begin{algorithm}[t]
\caption{Shared affordance-awareness for collaborative mobile robot manipulation of objects}
\label{alg:affordance_robot_alg}
    \textbf{Input}: Robot and human target objects, $\mathbf{o}_R, \mathbf{o}_H$; Human affordable area, $\mathcal{A}_H$; Detected objects, $\mathcal{O}$;\\
    \textbf{Output}: Collaborative robot plan;\\
    \textbf{Initialize}: Collision-free robot trajectory, $\tau_R$, towards target object $\mathbf{o}_R \rightarrow \tau_R$;
    
    \begin{algorithmic} %[1] enables line numbers
    \IF{$\tau_R$ found \AND $\mathbf{o}_R \not= \mathbf{o}_H$}
        \STATE \textbf{return} $\tau_R$
    \ELSIF{$\tau_R$ found \AND $\mathbf{o}_R == \mathbf{o}_H$}
        \STATE \textbf{re-plan} using new $\mathbf{o}_R$ from $\mathcal{O}$
    \ELSIF{$\tau_R$ not found \AND $\mathbf{o}_R \in \mathcal{A}_H$}
        \STATE query human to move $\mathbf{o}_R$ into robot affordable area
    \ELSIF{$\tau_R$ not found \AND $\mathbf{o}_R \notin  \mathcal{A}_H$}
        \STATE query human to see if $\mathbf{o}_R$ is reachable
    \ENDIF
    \end{algorithmic}
\end{algorithm}

\section{Proposed Experiment}

To evaluate the efficacy of the proposed affordance-aware framework, we consider an HRC experiment involving a domestic cleaning task. In this task setting, various objects from the YCB dataset~\cite{calli_ycb_2017} will be spread out among the floor, tables, and shelves (Fig.~\ref{fig:overview}). The task will require agents to collect and deposit these objects into one of three bins located around the environment, depending on the object's category: food items, kitchen items, and household tools. The varied arrangement of object starting locations and categories presents notable difficulties for an individual mobile manipulator, \textbf{or} a person with disabilities attempting to accomplish the task independently. 

In contrast with no communication and uni-directional robot-to-human communication, we hypothesize that our shared affordance-aware framework will:
\begin{itemize}
    \item \textbf{H1}: Demonstrate increased task efficiency, as measured by task completion time, intent switching frequency, and agent idle time.
    \vspace{-0.2cm}\item \textbf{H2}: Result in reduced mental workload for human participants, estimated by NASA-TLX~\cite{hart_nasa_1998} and quantitative metrics based on eye-related measures, e.g., pupil diameter~\cite{novak_workload_2014}.%intent switching, gaze tracking
    \vspace{-0.2cm}\item \textbf{H3}: Enhance usability ratings, as indicated by the System Usability Scale (SUS)~\cite{brooke_usability_1996}.
\end{itemize}

\subsection{Experimental Setup}

In our domestic cleaning task, a between-subjects study will be conducted with one of three experiment modes: non-communicative, robot-to-human communication only, or shared affordance-aware communication. In the non-communicative mode, the human and robot will act as independent agents with no means of sharing intent or affordances. In the second configuration, only the robot agent's intent and affordances will be displayed in the AR interface, as in previous work~\cite{quesada_proactive_2022}. In the final experiment mode, our framework will communicate intent, affordances, and allow agents to query one another for assistance. 

\subsection{Performance Metrics}

We will evaluate the performance of the collaborative human-robot team through subjective user responses and quantitative metrics. Using NASA-TLX~\cite{hart_nasa_1998} and SUS~\cite{brooke_usability_1996} scales, we will acquire participant opinions on our system's ease of use, helpfulness, perceived workload, as well as overall user experience. By equipping the AR headset with a Pupil Labs eye tracker add-on~\cite{kassner_pupil_2014}, we will record gaze fixation and pupil dilation signals during the task, hence providing suitable signifiers of the human agent's workload~\cite{lu_integrating_2020, kosch_look_2018, marquart_review_2015}. Lastly, we evaluate team efficiency through agent idle time (duration where an agent has no object of intent), frequency of intent switching, as well as time-to-completion. These metrics will provide a comprehensive overview of our framework's ability to facilitate effective HRC in a real-world collaborative task.

{\small
\bibliographystyle{ieee_fullname}
\bibliography{cvpr_references}
}
 
\end{document}